%% file: main.tex
\documentclass[10pt,twocolumn,letterpaper]{article}

\usepackage[pagenumbers]{cvpr} 

\usepackage{graphicx}
\usepackage{amsmath}
\usepackage{amssymb}
\usepackage{booktabs}

\usepackage{graphicx}
\usepackage{amsmath}
\usepackage{amssymb}
\usepackage{booktabs}
\usepackage{tabularx}
\usepackage{multirow}
\usepackage[]{algorithm2e}
\RestyleAlgo{ruled}

\newcolumntype{Y}{>{\centering\arraybackslash}X}

\include{preamble}

\newcommand{\js}[1]{{\color{black}#1}}

\usepackage[pagebackref,breaklinks,colorlinks]{hyperref}

\usepackage[capitalize]{cleveref}
\crefname{section}{Sec.}{Secs.}
\Crefname{section}{Section}{Sections}
\Crefname{table}{Table}{Tables}
\crefname{table}{Tab.}{Tabs.}

\usepackage{pifont}
\newcommand{\cmark}{\ding{51}}%
\newcommand{\xmark}{\ding{55}}%

\def\cvprPaperID{9926} 
\def\confName{CVPR}
\def\confYear{2023}
\begin{document}

\input{sec/0_metadata}
\maketitle
\vspace*{-15pt}


\def\thefootnote{*}\footnotetext{Equal contribution}\def\thefootnote{\arabic{footnote}}
\input{sec/0_abstract}
\input{sec/1_introduction}
\input{sec/2_related}

\input{sec/3_method}

\input{sec/4_results}
\input{sec/5_conclusions}

{
    \clearpage
    \small
    \bibliographystyle{ieee_fullname}
    \bibliography{macros,main}
}


\end{document}

%% file: preamble.tex
\usepackage{overpic}
\usepackage{enumitem} 
\usepackage{overpic} 
\usepackage{color}

\definecolor{turquoise}{cmyk}{0.65,0,0.1,0.3}
\definecolor{purple}{rgb}{0.65,0,0.65}
\definecolor{dark_green}{rgb}{0, 0.5, 0}
\definecolor{orange}{rgb}{0.8, 0.6, 0.2}
\definecolor{red}{rgb}{0.8, 0.2, 0.2}
\definecolor{darkred}{rgb}{0.6, 0.1, 0.05}
\definecolor{blueish}{rgb}{0.0, 0.3, .6}
\definecolor{light_gray}{rgb}{0.7, 0.7, .7}
\definecolor{pink}{rgb}{1, 0, 1}
\definecolor{greyblue}{rgb}{0.25, 0.25, 1}

\usepackage{blindtext}

\renewcommand{\paragraph}[1]{\vspace{1em}\noindent\textbf{#1}.}

\definecolor{rohitcolor}{RGB}{77, 157, 224}
\definecolor{nehalcolor}{RGB}{225, 85, 84}


%% file: sec/0_metadata.tex
\title{Beyond mAP: Towards better evaluation of instance segmentation }

\makeatletter
\renewcommand*{\@fnsymbol}[1]{\ifcase#1\or$\dagger$\else\@arabic{#1}\fi}
\makeatother

\author{
\begin{tabular}{cccc}
    Rohit Jena\textsuperscript{1}\thanks{Correspondence to: \texttt{rjena@seas.upenn.edu}}
    & Lukas Zhornyak$^*$ \textsuperscript{1} &  Nehal Doiphode$^*$ \textsuperscript{1}  &  Pratik Chaudhari \textsuperscript{1}\\
\end{tabular} \\
\begin{tabular}{ccc}
Vivek Buch \textsuperscript{2} & James Gee \textsuperscript{1} & Jianbo Shi \textsuperscript{1} \\
\end{tabular} \\
\begin{tabular}{cc}
\textsuperscript{1} University of Pennsylvania \hspace{2cm}& \hspace{2cm} \textsuperscript{2} Stanford University \\
\end{tabular} \\
{\small \texttt{\{rjena,zhornyak,lahen,pratikac,jshi\}@seas.upenn.edu, vpbuch@stanford.edu, gee@upenn.edu}}
}



%% file: sec/0_abstract.tex
\begin{abstract}

Correctness of 
instance segmentation constitutes counting the number of objects, correctly localizing all predictions and classifying each localized prediction.
Average Precision is the de-facto metric used to measure all these constituents of segmentation.
However, this metric does not penalize duplicate predictions in the high-recall range, and cannot 
distinguish 
instances that are localized correctly but categorized incorrectly.
This weakness has inadvertently led to network designs that 
achieve significant gains in AP but also introduce
a large number of false positives.
We therefore
cannot rely on AP to choose a model that provides an optimal tradeoff between false positives and high recall.
To resolve this dilemma, we review alternative metrics in the literature and propose two new measures to explicitly measure the amount of both spatial and categorical duplicate predictions.
We also propose a Semantic Sorting and NMS module to remove these duplicates based on a pixel occupancy matching scheme.
Experiments show that modern segmentation networks have significant gains in AP, but also contain a considerable amount of duplicates. 
Our Semantic Sorting and NMS can be added as a plug-and-play module to mitigate hedged predictions and preserve AP. 
\end{abstract}

%% file: sec/1_introduction.tex
\section{Introduction}
\label{sec:intro}

Tasks like classification and semantic segmentation have a fixed output space, i.e. the K-dimensional probability distribution of the classes and the per-pixel semantic class respectively.
For classification, we can use the zero-one loss, and for semantic segmentation we can use a per-pixel cross entropy loss.
On the other hand, instance segmentation is a challenging problem because the output is a set containing an arbitrary number of objects, and the network does not have knowledge of the number of objects in the scene \textit{apriori}.
Therefore, the model has to count the correct number of objects in the scene, localize them all and classify them correctly.
Deep learning for instance segmentation has two broad paradigms - top-down and bottom-up instance segmentation.
In bottom-up instance segmentation, the image is converted into per-pixel features, and pixel features are aggregated to predict objects. 
This is typically done by grouping or clustering the pixels based on some similarity in the feature space \cite{blendmask, Papandreou_2018_personlab, Zhou_2019_extremebottomup, Wang_2022_partaffinity, houghforest, watershed}.
In top-down instance segmentation, a model proposes a set of candidate proposals, out of which proposals not containing an object are removed.
This leaves us with a smaller set of proposals which are further passed into a localization and classification branch.
This is typically followed by an NMS step, since an object may have multiple candidate proposals, so duplicates must be removed.
\js{
Popular approaches are dominated by top-down methods where the network regresses a bounding box, mask, and category. 
Mask-RCNN \cite{maskrcnn, fastrcnn, liu2021swin} approaches it as a two-stage problem: localize the object, then predict the associated instance segmentation mask.
SOLO \cite{solo, solov2} builds on an anchor-free framework and directly regresses an object segmentation using a spatial grid feature as a probe.
More recent work based on Transformers (\cite{detr, instancequery}) explicitly learn a query in the network memory, then refines this prediction.
}
We can interpret all these top-down methods as 
implementing the query-key paradigm.
Each uses different \textit{query} designs: anchor box-based object proposal for Mask R-CNN, grid-cell for SOLO, or learnable latent features for DETR/QueryInst.
The Query-Key interaction aims to extract different representations of the object: ROI pooled features for MaskRCNN, center-based convolution filters for SOLO, and cross-attention features in DETR.

In analyzing why top-down methods consistently perform better than bottom-up methods, we make an unusual observation.
The qualitative performance of bottom-up methods is at par with that of top-down methods, but there is a significant gap in mAP.
Upon further analysis of the precision-recall curves in top-down methods, we find that mAP can be increased by increasing the number of low-confidence predictions.
We observe that recent design choices in the literature has exacerbated this problem.
In this work, we take a step back and analyze how mAP can be `gamed' by increasing false positives, explore other metrics in the literature, and propose metrics that explicitly quantify this amount of false positives, both spatially and categorically.
Furthermore, we propose a Semantic Sorting and NMS module to improve all metrics related to this excessive amount of prediction, only with a minimal dip in mAP.

\begin{figure}[t!]
    \centering
    \includegraphics[width=\linewidth]{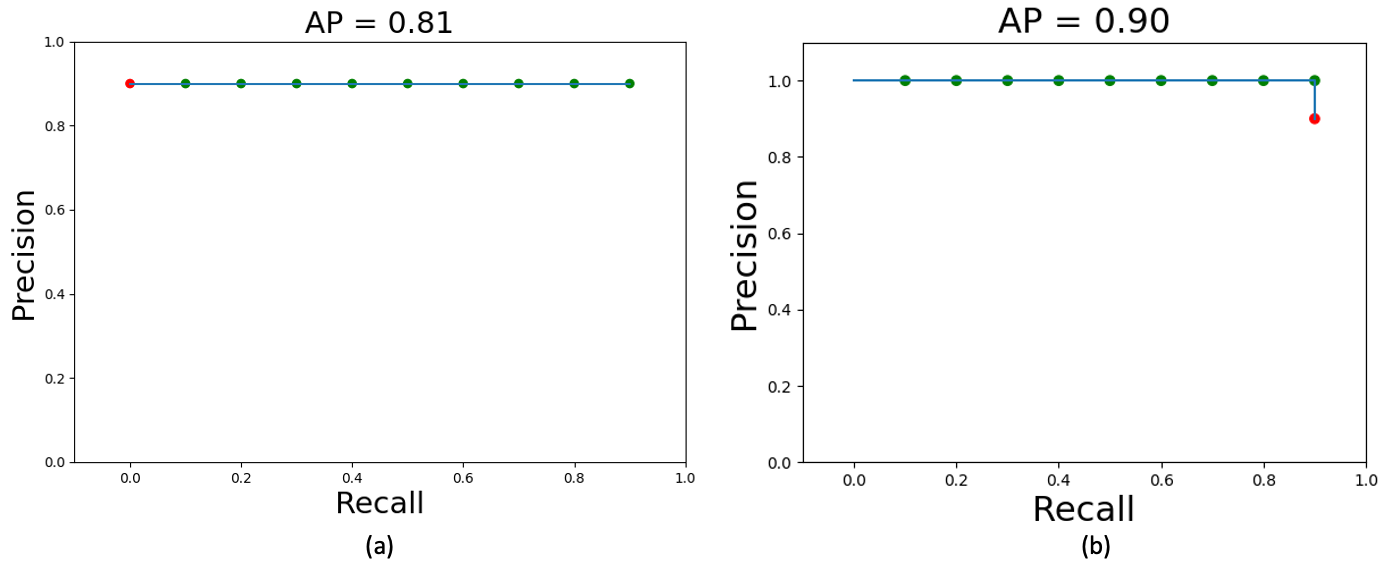}
    \includegraphics[width=\linewidth]{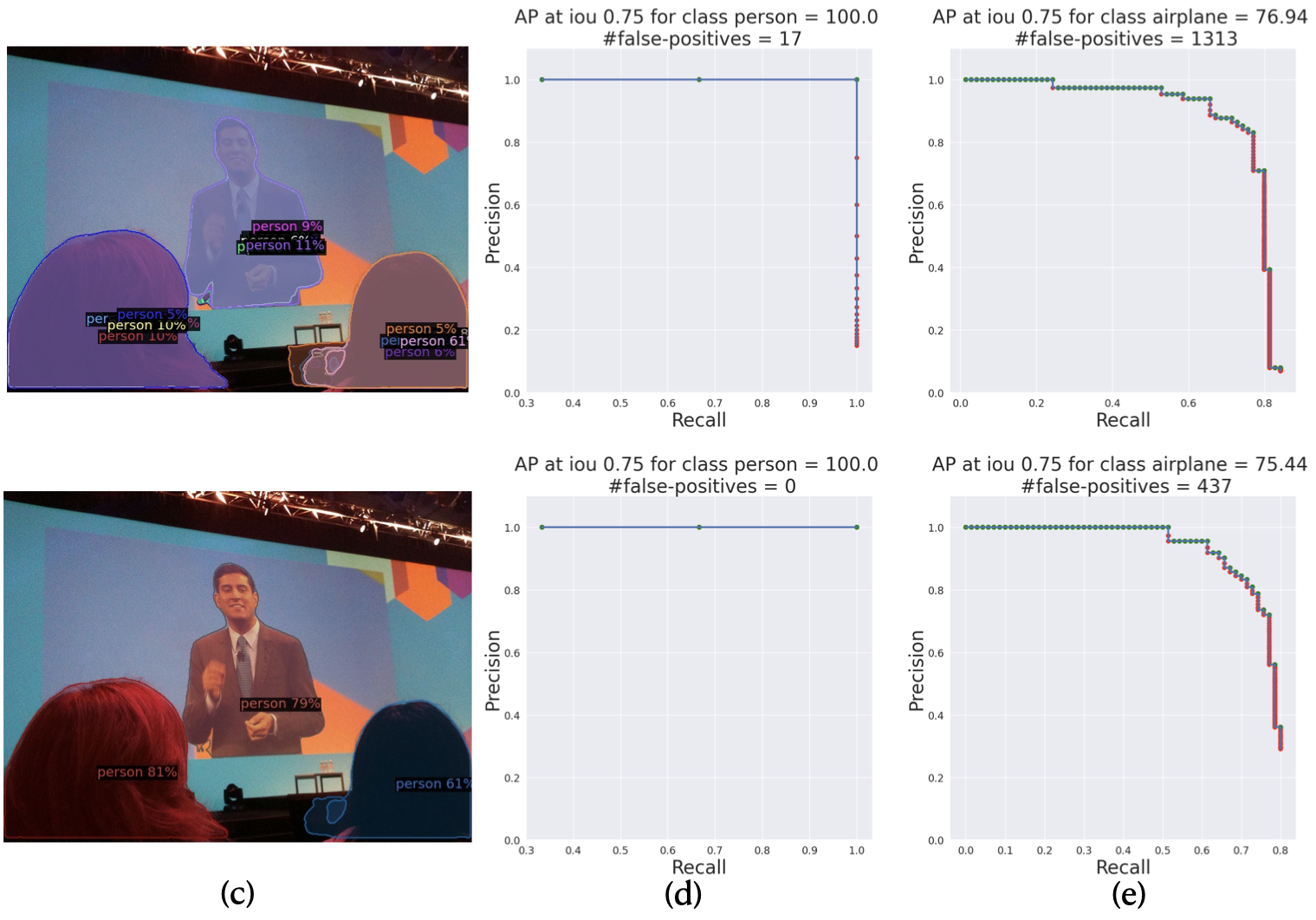}
    \caption{\textbf{Top}: Toy example demonstrating how AP changes with a reordering of the same set of detections (9 TPs, 1FP). 
    Note that in \textbf{(b)} the last FP doesn't contribute to AP. A detection that does not predict this example will also have the same AP.
    The last prediction in \textbf{(b)} is therefore a hedged prediction.
    \textbf{Middle, Bottom:} SOLOv2 with Matrix and Mask NMS respectively for the same network parameters.
    \textbf{(c)} shows the qualitative result and \textbf{(d)} is the corresponding P/R curve for the image. Note that hedged predictions do not penalize AP.
    \textbf{(e)} shows the P/R curve for \textit{airplane} category over entire COCO val dataset. Note that AP increases by 1 point, but number of false positives increase 3-fold. 
    }
    \label{fig:toyexample}
\end{figure}

%% file: sec/2_related.tex

\section{Related works}
\label{sec:related}
\js{
\subsection{Instance segmentation} 
Top-down instance segmentation is often viewed as a localization task followed by pixelwise classification of foreground masks.
Among such ``detect then segment'' strategies is FCIS \cite{fcis}, the first end-to-end fully convolutional work that considers position-sensitive score maps as mask proposals.
The score maps are then assembled to produce classification agnostic instance masks and category likelihoods. 
Along the same line of strategies is MaskRCNN \cite{maskrcnn}, a two-stage detector that predicts masks from proposed boxes after RoIAlign operation on feature maps.
YOLACT \cite{yolact} generates non-local prototype masks in an effort to learn and linearly combine them by predicting a set of mask coefficients.
However, it relies on accurate bounding box predictions, and doesn't learn to localize far-away instances.
BlendMask \cite{blendmask} attempts to combine FCIS \cite{fcis} and YOLACT \cite{yolact} in a hybrid approach.
Moving away from box-based object detection,  SOLO\cite{solov2} and CondInst\cite{condinst} take an anchor-free approach and use position-sensitive \textit{query} to extract object masks directly from the feature map.
All these approaches are driven in a top-down manner, where a few query points (often object centers) are responsible for predicting the whole object shape.
In contrast, bottom-up approaches focus on grouping pixels into an instance.
These approaches, including Hough-voting \cite{houghforest, implicitshape}, pixel affinity \cite{adaptiveaffinity, affinitycnn}, Watershed methods \cite{watershed}, pixel embedding \cite{assocembed, partspixels, recurrentinstgroup}, can be thought of as `flow' based: each pixel directly or indirectly learns to flow towards the object center.
This `flow' helps to group pixels to its object center, either in the image space or in a latent feature space, therefore localizing all objects simultaneously.
However, bottom-up methods are generally worse at localizing smaller objects, dealing with occlusion and crowded objects, and require complex aggregation and post-processing techniques ~\cite{blendmask, liu2018affinity}.
}

\begin{figure*}[t!]
    \centering
    \includegraphics[width=0.82\linewidth]{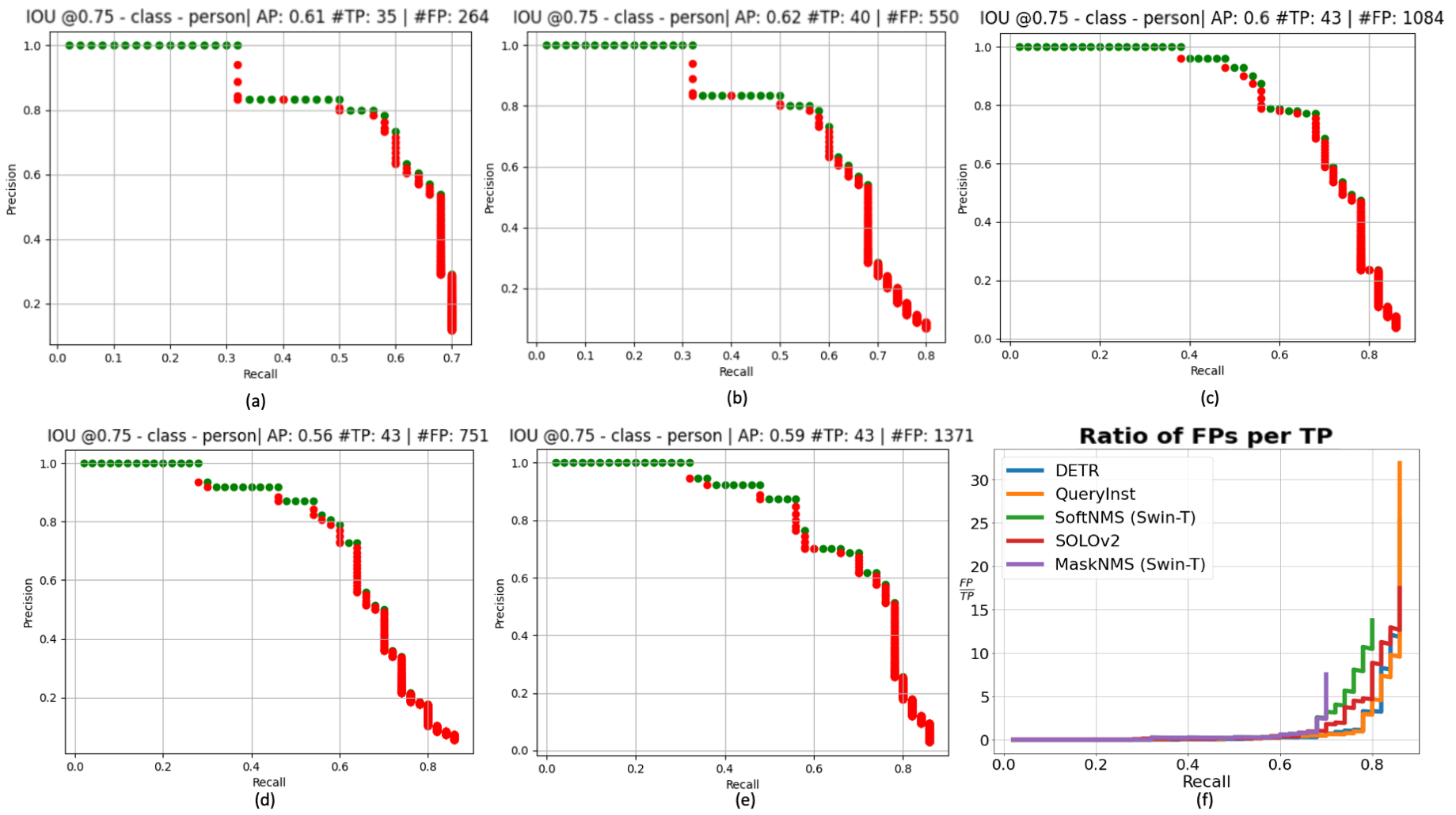}
    \caption{P/R curves for \textbf{(a)} Mask NMS, \textbf{(b)} SoftNMS, \textbf{(c)} SOLOv2, \textbf{(d)} DETR, and \textbf{(e)} QueryInst at IoU=0.75.
    Each point (red/green) on the P/R curve is a detection that is either a false positive or true positive respectively.
    \textbf{(f)} shows the ratio of FPs and TPs for (a-e). Note that modern segmentation networks have remarkably more hedging (as seen by the high FP:TP ratio).
    }
    \label{fig:apcurves}
\end{figure*}

\subsection{Evaluation of detection \& segmentation}
Average Precision (AP) \cite{everingham2010pascal} is the de-facto metric for measuring the performance of object detection and segmentation models.
Hoiem \etal \cite{hoiem2012diagnosing} provide a way to diagnose the effects of false positives and how they can be mitigated to improve mAP.
TIDE \cite{bolya2020tide} also provides a toolkit to identify and decompose the error (1 - mAP) into its constituent error components - such as classification, localization, duplication errors.
This can allow a researcher to analyse the major shortcomings of a given detection and segmentation method.
AP is therefore widely accepted in the community, and has remained unchallenged as a measure.
However, recent works have pointed out different shortcomings in mAP as a reliable metric.
Dave \etal \cite{dave2021evaluating} show that in a large vocabulary detection task, it is possible to gamify the AP metric by adding nonsensical predictions to a given prediction model.
LRP \cite{lrp} highlights two major problems with mAP: 1) different detectors having different P/R curves can have similar APs, but they have different underlying shortcomings, and 2) mAP is not sufficient to quantify localization.
LRP also acts as a desirable performance measure in terms of setting an optical confidence score threshold per class, unlike mAP which is optimal at a confidence threshold of 0 for any given model.
Our work is similar to \cite{dave2021evaluating}, but we show that AP can be `gamed' by adding low-confidence false positives, even with a moderate vocabulary task like COCO \cite{coco}.
This deficiency of mAP has led to design choices that inadvertently increase mAP by exploiting this behavior \cite{detr,solov2,instancequery}.
We capture aspects of this deficiency very explicitly, how they influence AP, quantify it and propose a plug-and-play module to mitigate this problem.

%% file: sec/3_method.tex
\input{sec/method-subsec/3_method_ap}

\input{sec/method-subsec/3_method_measuring}

\input{sec/method-subsec/3_method_semseg}

\subsection{Implementation details}
We run and evaluate all the baselines in the MMDet \cite{mmdetection} and Detectron2 \cite{wu2019detectron2} frameworks.
We implement our evaluation metrics within the \textit{pycocotools} framework \cite{pycocotools} to enable rapid adoption of the metrics into currently existing object detection and segmentation frameworks.

%% file: sec/method-subsec/3_method_ap.tex

\section{Preliminaries}
\label{sec:prelim}
Average Precision (AP) measures the area under the precision-recall curve of the detector, where the predictions over the dataset are sorted by confidence.
See ~\cite{pycocotools,everingham2010pascal} for details on how AP is computed.
LRP \cite{lrp} points out that AP is not confidence-score sensitive.
It is instead rank-sensitive.

Consider the following scenario.
A detector has a set of 10 predictions for a category with 10 ground truths. 
Consider two cases where one of the predictions is a false positive (FP).
If the highest confidence prediction is a FP, then the maximum precision is 0.9, and the AP is \textit{0.81}.
If the lowest confidence prediction is a FP, then the precision stays at 1 throughout recall 0 to 0.9, leading to AP of \textit{0.9}.
The AP curves are shown in Fig.\ref{fig:toyexample}(top).
This is a desirable property in a measure, to penalize a higher confidence FP more than a lower confidence FP.  
However, note that in the second case, the last 
FP
did not contribute (negatively) to the AP. 
This is the basis of the shortcoming that we highlight.

\subsection{Hedged predictions}
\label{sec:hedging}
In this section, we introduce the terminology around \textit{hedged predictions} in segmentation.
Hedging occurs when a segmentation framework outputs a lot of low-confidence \textit{duplicate} predictions, which are 
marginally different versions
of an `original' high confidence prediction.
This perturbation is both spatial and across object categories.
We call these predictions \textit{hedged predictions} because the network hedges these low-confidence predictions to increase mAP.
Low-confidence FPs do not affect AP, but TPs in this low-confidence, high-recall regime can boost AP slightly.

\paragraph{Spatial Hedging} \textbf{(SH)} refers to hedged predictions which are spatially perturbed versions of each other.
Spatial Hedging is done by design in top-down detection and segmentation frameworks.
In Mask-RCNN \cite{fastrcnn,maskrcnn} the locations in the region proposal network (RPN) are trained for objectness by imposing a minimum IoU criteria with ground truth objects.
A single ground-truth box/mask may assign positive labels to multiple anchors.
This leads to a hedging scheme where nearby anchors are incentivized to predict the same object.
In SOLO \cite{solo,solov2}, a grid cell is labeled as a positive  grid if it falls in the \textit{central region} of any ground truth object.
As such, multiple grid cells may be assigned to predict the same ground truth object.
HTC \cite{htc} uses the Mask-RCNN framework as a backbone, and shares the same RPN setup as MaskRCNN.
FCOS \cite{tian2019fcos} also assigns positive labels to spatial locations which fall into any ground truth box, which is used in PolarMask \cite{xie2020polarmask}.

All these are examples of spatial hedging where neighboring spatial anchors are encouraged to predict the same object.
Spatial hedging occurs to increase recall, in case the `central' anchor fails to predict the object. 
Typically, NMS is employed to discard duplicate predictions.
Recent works \cite{softnms,detr,solov2} have proposed alternate NMS that are faster than Mask NMS, and produce better mAP.
A closer inspection reveals that the methods are very effective in decaying the confidence of duplicate predictions, but the post-NMS confidence threshold is chosen to be too low (as low as 0.05 for SOLOv2 and 0.001 for SoftNMS) which leads to lot of duplicates retained post-NMS contributing to hedging.

\paragraph{Category Hedging} \textbf{(CH)} occurs when a model predicts multiple object categories for a single object detection.
This is a more subtle hedging, which 
seems to occur due to the way we do inference 
\ie choosing top-k classes for an object mask \cite{instancequery}, or selecting all classes that cross a particular threshold \cite{solov2}.
Note that selecting top-k classes during inference is inconsistent with the instance segmentation task - an object can only be of a single class.
The other k-1 classes then end up being hedged predictions.
Traditional NMS cannot mitigate this kind of hedging because NMS methods are class-independent.
If there are no competing objects in the hedged categories (a car which is also hedged as a truck), then the hedged object will not be removed by NMS.
This motivates the need for a category-aware NMS to suppress duplicate objects.
This forms the basis for our Semantic NMS in Sec.\ref{sec:semanticnms}.

\begin{figure}[t!]
    \centering
    \includegraphics[width=\linewidth]{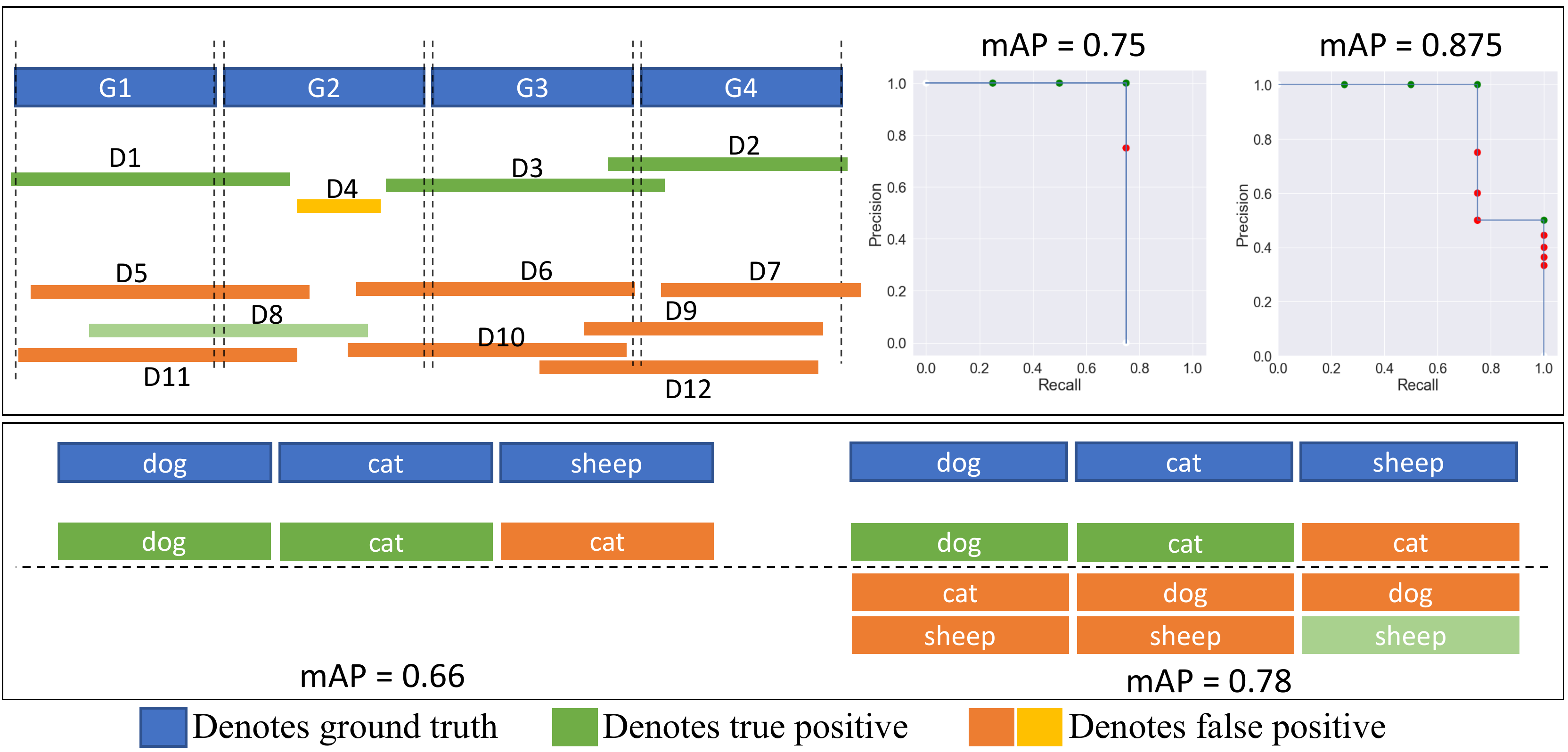}
    \caption{\textbf{Illustration of spatial and categorical hedging}: \textbf{Top:} Given ground truths G1-G4 (blue), predictions D1-D4 produce an AP of 0.75 (D4 is a FP).
    Spatially perturbing D1-D4 to produce objects D5-D12 results in a match of D8 with G2, leading to AP of 0.875.
    }
    \label{fig:hedging}
\end{figure}

Methods that are based on a fixed number of queries \cite{detr,instancequery} can perform both spatial and category hedging.
QueryInst\cite{instancequery} and DETR\cite{detr} use a fixed number of object proposals per image, and a top-k category selection for each proposal can lead to both forms of hedging.
DETR explicitly mentions that for all proposals where the top class prediction was `background', replacing the class of those proposals to the second-most confident class increased mAP by 2 points.
However, this change essentially predicts 100 objects for \textit{each} image, showing that explicit hedging does improve mAP.
However, from an end users' perspective 
this is not desirable,
since the user may be more satisfied with a better FP-FN tradeoff than predicting a 100 objects for each image.
We take a closer look at the P/R curves of pretrained models that are used to evaluate mAP. 
Indeed, there is a tremendous amount of hedging that show up as false positives in the P/R curves.
We show this in more detail in Fig. \ref{fig:apcurves}.
We use the Swin-T model with NMS and SoftNMS in Fig.\ref{fig:apcurves}(a,b) with a post-NMS threshold of 0.01 for SoftNMS. This is much higher than the default threshold of 0.001.
Comparing Fig.\ref{fig:apcurves}(a) and (b), we see that 5 new TPs are added, but they are all in the $\ge$0.7 recall range.
Moreover, we add 286 false positives to add only 5 true positives (57.2 FPs added per TP).
However, AP doesn't penalize these newly added FPs. 
This problem also occurs in state-of-the-art frameworks \ie DETR, SOLOv2, QueryInst (Fig.\ref{fig:apcurves}(c-e)) where the FPs added per TP rate is upto 138x compared to (a).
In Fig.~\ref{fig:apcurves}(f), we plot the ratio of FPs and TPs with recall.
Other methods are effective at having a lower FP/TP ratio at a medium recall range, but it quickly increases in the high recall range due to hedged predictions.
Moreover, the AP values are very similar, indicating that hedging is not penalized by AP.
More results are present in the Appendix.
This motivates the need to quantify the amount of hedging in a segmentation framework. 

%% file: sec/method-subsec/3_method_measuring.tex
\section{Measuring hedging}
\label{sec:measuring}
We propose to quantify the amount of both spatial and categorical hedged predictions. 
The main idea we use to measure spatial hedging is to measure the amount of mask overlap between predicted instances, weighed by their confidence scores. 
For category hedging, the main idea is to match predictions and ground truth objects based on mask overlap alone, and then measure the mismatch in categories of matches.

\begin{figure}[h]
    \centering
    \includegraphics[width=\linewidth]{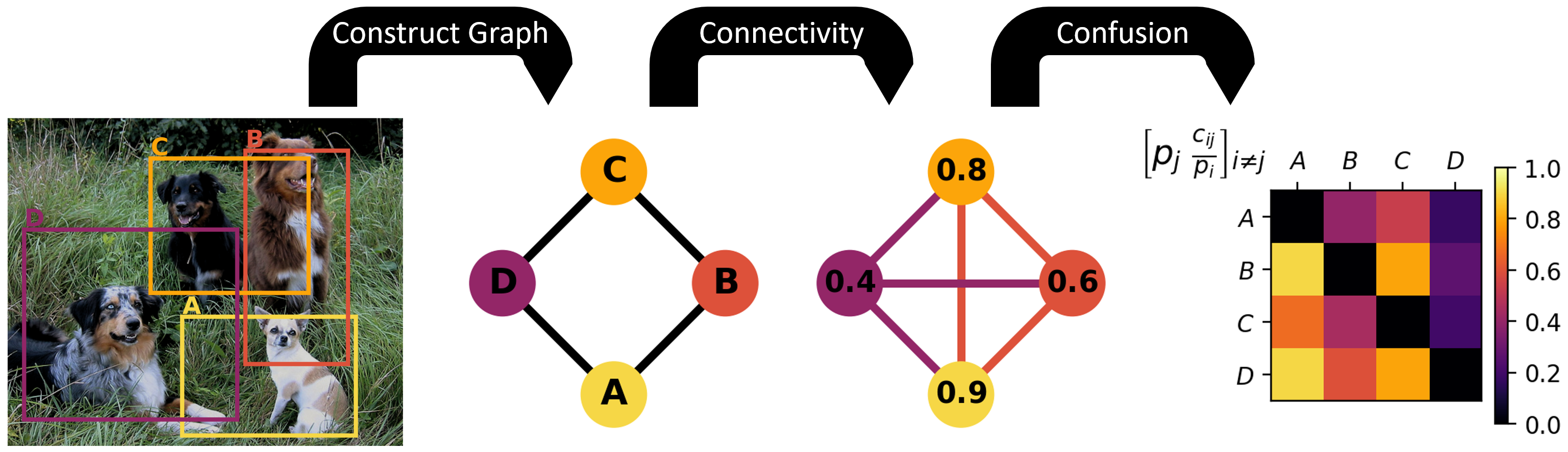}
    \caption{Illustration of duplicate confusion: It works by constructing a graph of objects based on IoU and then measuring a `max-flow-like' measure shown in Eq.\ref{eq:confusion}.}
    \label{fig:dc}
\end{figure}

\subsection{Duplicate Confusion Error}
\label{sec:dcerror}
Duplicate confusion (DC) measures how connected a given prediction is to all other predictions in terms of overlap, and aggregates this quantity for all predictions.
To do this, we first consider an IoU threshold $t$ and confidence threshold $v$ and object category $k$.
Now, consider a set of detections $\{d_1, d_2 \ldots d_m\}$ of category $k$ for image $I$. 
Let prediction $d_i$ have a confidence score of $\tau_i$.
We construct a graph $\mathcal{G} = \{ \mathcal{V}, \mathcal{E} \}$ where $\mathcal{V} = \{d_1 \ldots d_m\}$ is the set of predictions and $\mathcal{E} = \{(i, j): \text{IoU}(d_i, d_j) \ge t \}$ is the set of edges between these vertices.
The connectivity of the graph indicates predictions that have an IoU of at least $t$.
With slight abuse of notation, we refer to node $i$ in graph $\mathcal{G}$ to denote detection $d_i$.
For two nodes $i \ne j \in \mathcal{G}$, consider a path $\pi_{ij} = [u_1 = i, u_2, \ldots u_M = j]$ such that $(u_a, u_{a+1}) \in \mathcal{E} \forall a$ and $a \ne b \implies u_a \ne u_b$.
Consider the set of all such paths $T_{ij} = \{\pi_{ij}\}_{\pi}$.
We define the `connectivity' of $i$ and $j$ as 
\begin{equation}\label{eq:connectivity} \small
    c_{ij} = \max_{\pi \in T_{ij}} \min_{k \in \pi} \tau_k 
\end{equation}
Roughly, the inner min calculates the `weakest link' in terms of confidence score in the path $\pi$, which bottlenecks the (indirect) overlap between $i$ and $j$.
The max calculates the highest overlap among all such paths.
We solve a variant of the \textit{max-flow} problem to characterize the overlap of predictions $i$ and $j$ where the flow along the path $\pi$ is bottlenecked by $\min_{k \in \pi} \tau_k$.
If there are no paths between $i$ and $j$, then $c_{ij} = 0$.
Now, the total amount of overlap for a prediction $i$ is simply equal to 
\begin{equation}
    \label{eq:totalflow} \small
    \sum_{j \ne i}\tau_j c_{ij}
\end{equation}
which is the weighted sum of max-flow to all other nodes $j$ weighted by confidence $\tau_j$.
However, for a lot of low-confidence false positives, $c_{ij}$ will have very small absolute values and will be overshadowed by few high-confidence predictions.
To alleviate this problem, we perform relative scaling on Eq.\ref{eq:totalflow} by weighing by $1/\tau_i$.
We then aggregate this over all detections $d_i$ to get the final connectivity score:
\begin{equation}\label{eq:confusion}
    \mathrm{DC}_{tv} = \frac{1}{m} \sum_i^m \sum_{j \ne i} \frac{\tau_j c_{ij}}{\tau_i}
\end{equation}
Similar to mAP, we compute DC by averaging over IoU thresholds $t$ and confidence thresholds $v$.

This can be interpreted as the confidence of a network in its own counting.
This is not, however, a measure of how effectively a network can count instances - the ground truth is not considered when calculating DC.
It is important to emphasize that DC by itself is not a good measure, \ie producing no predictions trivially results in 0 DC.
This is analogous to model calibration error ~\cite{guocal}, where 0 calibration error can be achieved trivially by predicting the overall label distribution for every input. 
However, DC captures spatial hedging effectively, since spatially perturbed predictions will form densely connected graphs, therefore accounting for a quadratic number of interactions among these hedged predictions.

\begin{figure}[t!]
    \centering
    \includegraphics[width=\linewidth]{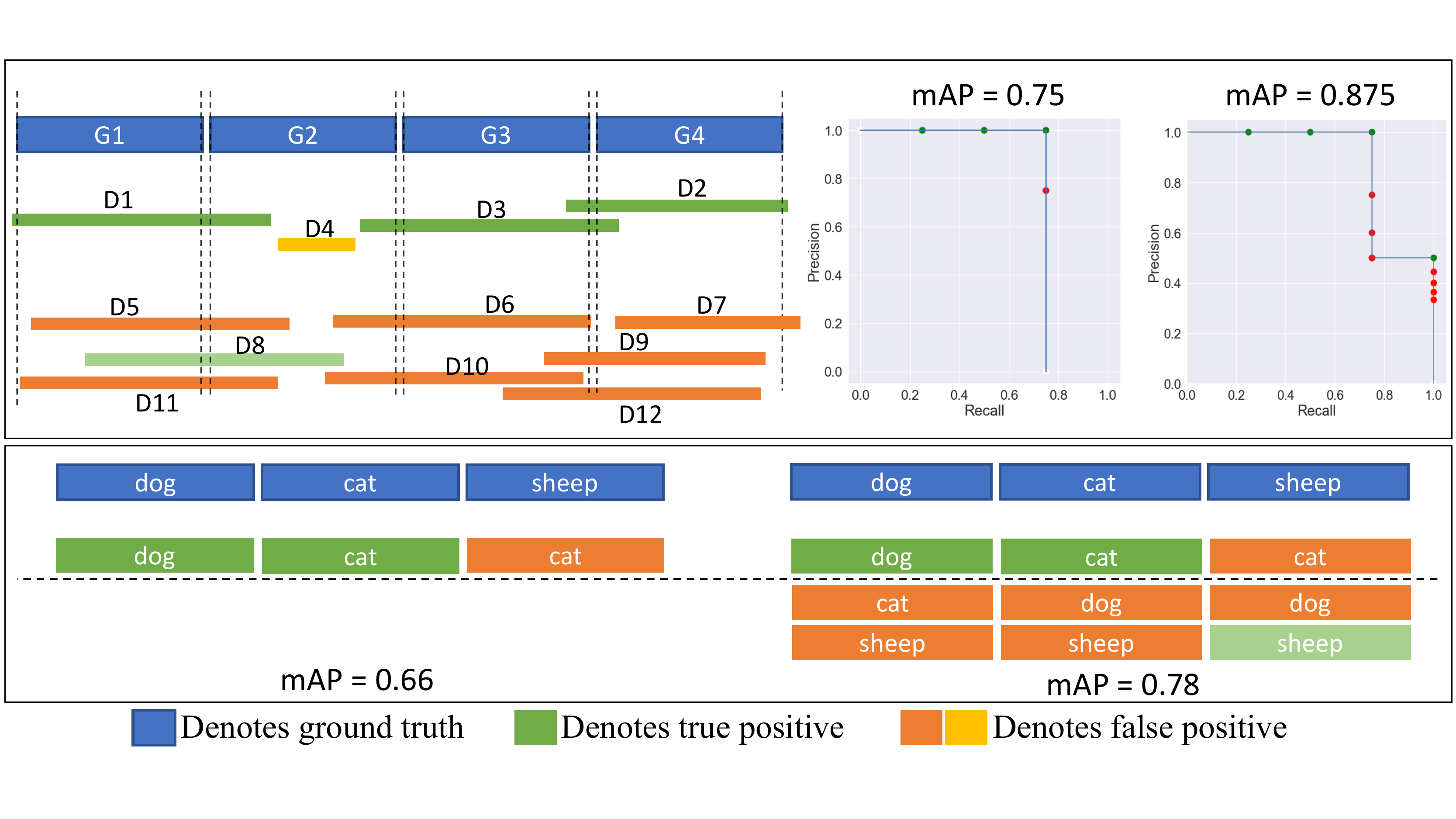}
    \caption{\textbf{Illustration of Naming Error (NE):} GTs and detections are matched in a category-agnostic manner, and then the labels are revealed. The average number of mismatches per GT is the NE.}
    \label{fig:namingerror}
\end{figure}

\subsection{Naming Error}
\label{sec:neerror}
Measuring category hedging requires us to examine correctly localized object detections and analyse if 
their category is predicted correctly.
Category hedging occurs when the model hedges on the category of the object with a single mask \ie top-k category predictions.
To measure category hedging, we need to treat object detections and ground truths in a category-agnostic manner.

Let $\mathcal{G} = \{G_1, G_2 \ldots G_n\}$ be the set of \textit{all} ground truth objects in image $I$ and $l_{G_i}$ be the category of ground truth $G_i$.
Similarly, let $\mathcal{D} = \{ D_1, D_2 \ldots D_m\}$ be the set of \textit{all} detections output from the network, and $l_{D_j}$ be the category of detection $D_j$.
The key intuition we use is that if detection $D \in \mathcal{D}$ is a category hedged prediction, it will have a high mask overlap with some ground truth object $G \in \mathcal{G}$, but will have a label mismatch.
For each detection $D_j$, we define its ground truth match as:
\begin{equation}
\begin{small}
    g(D_j) = \begin{cases}
        \arg \max_i\text{IoU}(D_j, G_i) & ,\max_i \text{IoU}(D_j, G_i) \ge 0.5 \\
        -1 & , \text{otherwise}
    \end{cases}
\end{small}
\end{equation}
This matches each detection to a ground truth object or no object depending on its maximum overlap with the set of ground truth objects.
Using the function $g$ we define the \textit{Naming Error} as:
\begin{equation}
    \mathrm{NE} = \frac{1}{N} \sum_{i=1}^N \sum_{j: g(D_j) = i} \mathbb{I}\left[ l_{D_j} \ne l_{G_i} \right]
\end{equation}
This gives us the average number of detections that have mismatched labels with a ground truth object.

%% file: sec/method-subsec/3_method_semseg.tex
\section{Semantic Sorting and NMS}
\label{sec:semanticnms}

\begin{algorithm}[t!]
\caption{Pseudocode for semantic sorting and NMS, given instances $D_k$ with category $c_k$ and confidence $\tau_k$, threshold $thr$, semantic masks $M$}\label{alg:semnms}
\KwData{$\{ D_{k}, c_{k}, \tau_{k}\}_{k = 1\ldots N}$, $\{ M_c \}_{c = 1\ldots C}$}
\KwResult{Boolean array $keep$ indicating preservation of instances}
\For{$k = 1 \ldots N$} {
    $pr \leftarrow \text{precision}(D_k, M_{c_k} ) $; \\
    $iou \leftarrow \text{IoU}(D_k, M_{c_k} ) $; \\
    $\tau_k \leftarrow \tau_k + pr + (1 - iou) $; \\
}
$(D, c, \tau) = \text{sort}(D, c, \tau); \quad$ // sort by decreasing $\tau$ \\
\For{$k = 1 \ldots N$}{
    $overlap \leftarrow \text{precision}(D_k, M_{c_k})$;  \\
    \eIf{$overlap \ge thr$}{
        $keep[k] = True$; \\
        $M_{c_k} = M_{c_k} \backslash D_k$ \\
    }{ $keep[k] = False$}
}
\end{algorithm}

Techniques like NMS have been designed to resolve hedging in principle.
However, newer NMS designs \cite{softnms,detr,solov2} 
use ``softer'' penalties on duplicate predictions to decay their confidence scores based on overlap with other detections, and threshold the confidence score.
Although these designs have led to significant improvements in mAP, we observe that they also contribute to the hedging problem.
SoftNMS and SOLOv2 choose a very low post-NMS threshold (0.001 and 0.05 respectively), and DETR selects the next-best class for background predictions.
We notice that these design choices lead to considerable spatial hedging (see Fig.\ref{fig:apcurves}).
Moreover, we notice that NMS does not consider suppression between different categories, which is important to mitigate category hedging.
This is difficult to design based on detection masks and confidence scores alone, since neural networks are often miscalibrated ~\cite{guocal}, especially for long-tailed segmentation tasks ~\cite{pan2021model, dave2021evaluating, Tan_2020_equalization}.

\paragraph{Semantic Sorting}  
To alleviate this, we propose a Semantic Sorting algorithm, which re-scores each instance based on its `agreement' with a semantic segmentation output, and a Semantic NMS which discards an instance if its `mask occupancy' is not supported by the semantic mask.
Given an image, we predict a semantic segmentation mask $M$.
This can either be predicted from an off-the-shelf network, or can be learnt alongside instance segmentation \cite{htc}.
For each instance $D_k$ with category $c_k$ and confidence $\tau_k$, we measure the fraction of pixels in $D_k$ that match with semantic mask $M_{c_k}$ (precision of $D_k$).
A low precision implies a lower-quality mask which doesn't overlap well with the semantic segmentation.
We also compute the IoU of the detection and semantic mask, and favor instances with smaller IoU, to retain smaller objects in case of crowded instances.
These scores are added to $\tau_k$ and averaged. 
This score is then used to reorder the detections for Semantic NMS (Alg.\ref{alg:semnms}).

\begin{table}[t!]
\centering
    \begin{tabularx}{\linewidth}{lccccc}
    \toprule
    \textbf{Model} & \textbf{CoordConv} & \textbf{AP\tiny{50}} & \textbf{F1\tiny{0.5}} & \textbf{LRP} & \textbf{LRP\tiny{Loc}}\\
    \midrule
    SOLOv2 & \xmark & 96.87	& \underline{0.47} & 79.65 & 16.55 \\
    SOLOv2 & \cmark & \underline{96.90} & 0.46 & 79.87 & 16.06 \\
    Ours & \xmark & \textbf{98.01} & \textbf{0.99} & \underline{33.46} & \underline{15.87} \\
    Ours & \cmark & \textbf{98.01}	& \textbf{0.99}	& \textbf{33.37}	& \textbf{15.75} \\
    \bottomrule 
    \end{tabularx}
    \includegraphics[width=\linewidth]{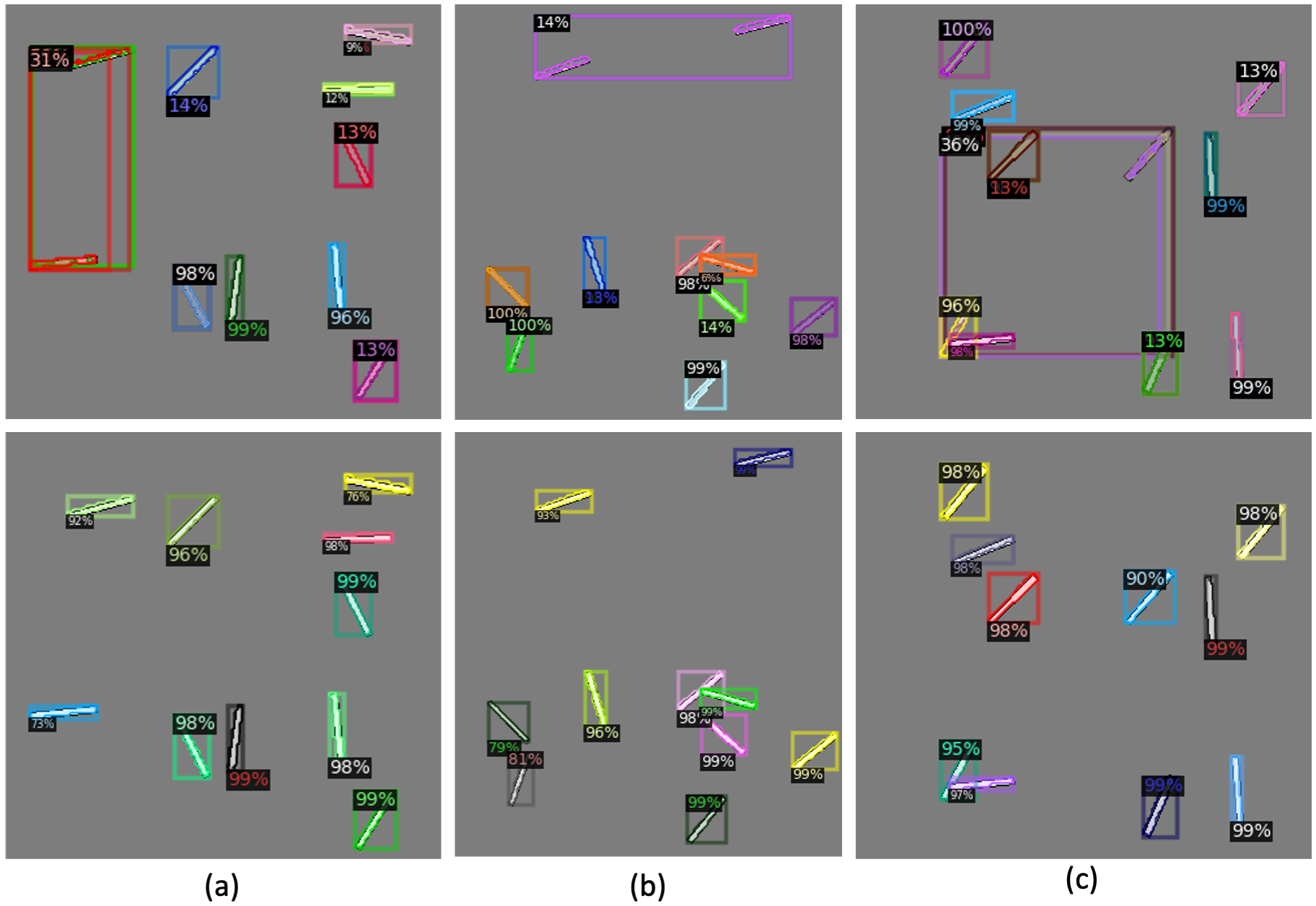}
    \caption{Results on synthetic part counting dataset.
    Note that Semantic Sorting and NMS effectively resolves spatial hedging (as seen by the tremendous improvement in F1 and LRP).
    Figure (top) shows predictions by SOLOv2 and (bottom) shows our method.
    }
    \label{tab:synthetic} 
\end{table}

\input{tab/ablation}

\paragraph{Semantic NMS}
To alleviate both spatial and category hedging, we propose the use of Semantic NMS.
The key idea here is that NMS discards objects based on a minimum IoU threshold with each other.
However, we treat NMS as an occupancy problem \ie an instance whose occupancy is not supported by the semantic mask is discarded.
We do this by calculating the fraction of the pixels in $D_k$ that are present in $M_{c_k}$.
A substantial overlap indicates that the detection with its category $c_k$ has occupancy in $M_{c_k}$ and the detection is preserved.
We update $M_{c_k}$ by subtracting the mask $D_k$, deleting the occupancy of this region.
Another object with a similar mask will have low overlap with this new $M_{c_k}$ and will be discarded.
This NMS provides us two other benefits: (1) it considers the predicted category to determine whether to keep or discard instances, (2) it runs in $O(n)$ time unlike NMS which runs in $O(n^2)$ time.


%% file: tab/ablation.tex
\begin{table*}[ht!]
\centering
\includegraphics[width=0.8\linewidth]{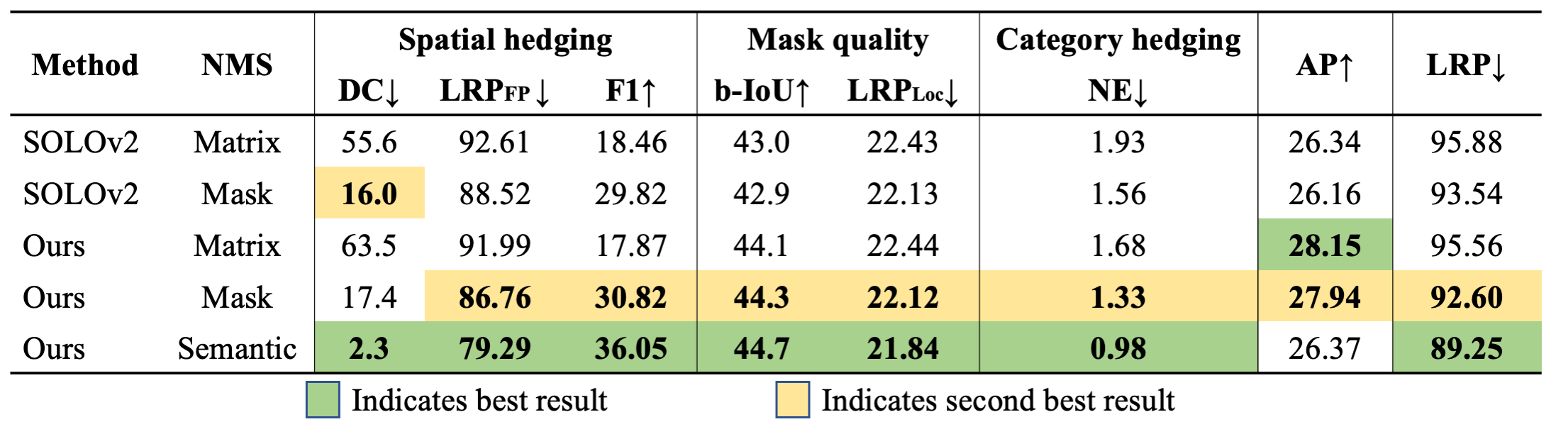}
\caption{
\textbf{Ablation of NMS with SOLOv2}: We perform ablations on the type of NMS on coco-minitrain \cite{houghnet} dataset. 
Adding the semantic sorting and NMS modules leads to better resolution of hedging errors, both spatial and categorical.
The only caveat is a slight drop in AP, which comes from pruning the long-tail of the P/R curve, which is generally acceptable from an end-user's perspective.
} 
\label{tab:ablation}
\end{table*}

%% file: sec/4_results.tex
\section{Experiments}
We first analyze the effectiveness of Semantic Sorting and NMS on a part-counting toy problem, perform an ablation of different NMS by using SOLOv2 as an illustrative example, and compare different state-of-the-art architectures on their performance of mAP, LRP, and our proposed measures.
We also compare boundary IoU \cite{Perazzi_2016_CVPR} and LRP$_{Loc}$ for localization quality, and F1-score of detections for an overall hedging measurement 
(F1 uses precision, which will be low with many hedged predictions). 
Unless otherwise mentioned, we use SOLOv2 as the base framework on which we implement our modules.

\input{tab/cocoresult}

\vspace*{-2pt}
\subsection{Part counting dataset}
To isolate the \textit{spatial hedging} problem, we construct a synthetic part counting dataset.
Each image has 10 identical nails that are placed randomly in the image.
The locations of the nails are sampled from a truncated random normal distribution around the image center and the nails are placed sequentially.
Therefore, nails may be occluding each other, and the ground truth masks are constructed accordingly. 
Since there is only one class, we can isolate the effect of spatial hedging.
Results are in Table.\ref{tab:synthetic}.
Note that SOLOv2's over-reliance on appearance-based features in its kernel branch leads to instances that are pooled together based on convolution of a kernel feature with the entire mask feature.
CoordConv doesn't seem to resolve this issue either.
Moreover, the AP is very high, which may mislead an user to think that the model is very strong.
Our semantic sorting + NMS leads to drastic improvements in F1 score and LRP, showing better resolution of hedging.

\subsection{Ablation on coco-minitrain}
We perform ablations on the coco-minitrain \cite{houghnet} dataset.
We use coco-minitrain instead of the COCO-train-2017 set owing to similar data statistics as the full training set and to reduce the cost of running ablations.
All hyperparameters used for SOLOv2 follow the experimental setup of \cite{solov2}.
The results are in {Table \ref{tab:ablation}}.
Overall, using semantic NMS provide at least a 86.8 $\%$ decrease in the duplicate confusion and a 15.4 $\%$ increase in the F1 score compared to Matrix and Mask NMS. 
Using Semantic NMS leads to a much better DC, F1, LRP$_{FP}$, and NE showing better resolution of both 
spatial and category hedging.

\begin{figure*}[ht!]
    \centering
    \includegraphics[width=0.9\linewidth]{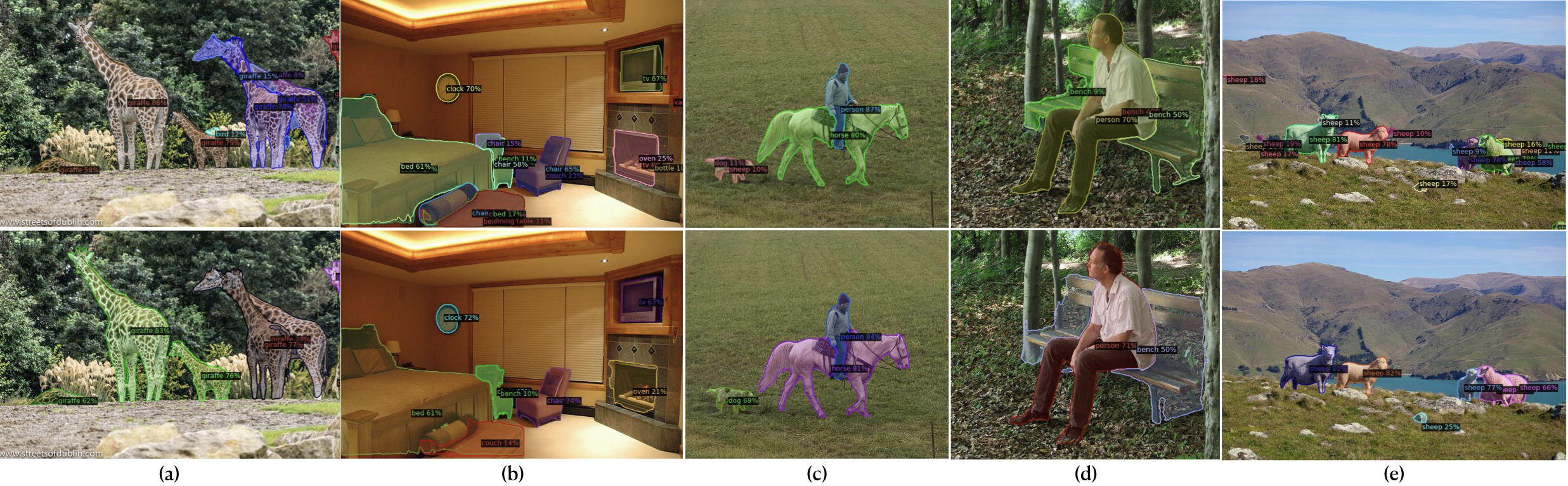}
    \caption{\textbf{Qualitative results on COCO val:} Top (SOLOv2) and bottom (Ours) show surprisingly different results with only a minimal difference in mAP.
    In (b,c) our method removes category hedging (chair $\rightarrow$ couch, bed $\rightarrow$ bench, couch $\rightarrow$ dining table, dog $\rightarrow$ sheep) and removes spatial hedging in (a,d,e).
    This drastic difference in hedging is reflected in F1 score, LRP, DC and NE measures (Table~\ref{tab:cocoresult}).
    }
    \label{fig:qualexamples}
\end{figure*}

\subsection{Performance on COCOval dataset}
We run our method on the COCO \cite{coco} training set. The results are in Table \ref{tab:cocoresult}.
We perform comparison with several SOTA methods to contrast the effect of Semantic NMS with Matrix NMS.
Methods like QueryInst \cite{instancequery} use a fixed number of queries (\ie 100) and produce predictions for each query.
For R101, QueryInst has the highest AP value, but has the poorest performance in terms of LRP, F1, LRP$_{FP}$ and NE, showing that its predictions are prone to both spatial and category hedging.
Modern instance segmentation methods perform very competitively in terms of AP, but also produce a noticeable quantity of hedged predictions. 
MaskRCNN has relatively lower spatial hedging because Mask NMS is effective at removing duplicates , albeit at a higher computational cost.
MaskRCNN also uses one category prediction per mask, therefore has the lowest category hedging as well.
Our method is built on SOLOv2, and we observe upto a 33x improvement in DC, and upto 10 points of LRP and 19 points of LRP$_{FP}$ showing significant resolution of spatial hedging.
Notice that the mask quality of our final detections is not compromised, as shown by b-IoU and LRP$_{Loc}$ values.
There is a slight drop in AP, which occurs because Semantic NMS essentially prunes the `tail-end' of the P/R curve to delete hedged predictions.


\vspace*{-2pt}
\subsection{Speedup}
Semantic NMS has a $O(n)$ running time complexity, making it faster than Mask NMS while improving all metrics quantifying hedging including LRP, F1, DC, NE and a slight improvement in boundary IoU.
On coco-minitrain, Semantic NMS needs an average of $0.028 \pm 0.027$ seconds, while Mask NMS needs an average of $0.169 \pm 0.290$ seconds, achieving $\sim6.03\times$ speedup.
Our NMS can thus prove beneficial when the number of objects is large, such as crowd segmentation.

\vspace*{-2pt}
\subsection{Qualitative results}
Finally, we analyse some qualitative examples comparing a SOLOv2 base model and SOLOv2 with semantic sorting and NMS.
Outputs are shown without any other post-processing, apart from that performed by the NMS.
In Fig.\ref{fig:qualexamples} we note that our method resolves both spatial and category hedging, 
while Mask NMS can only resolve spatial hedging.
More qualitative results are in the Appendix.

%% file: tab/cocoresult.tex
\begin{table*}[t!]
\small
\centering
\includegraphics[width=0.82\linewidth]{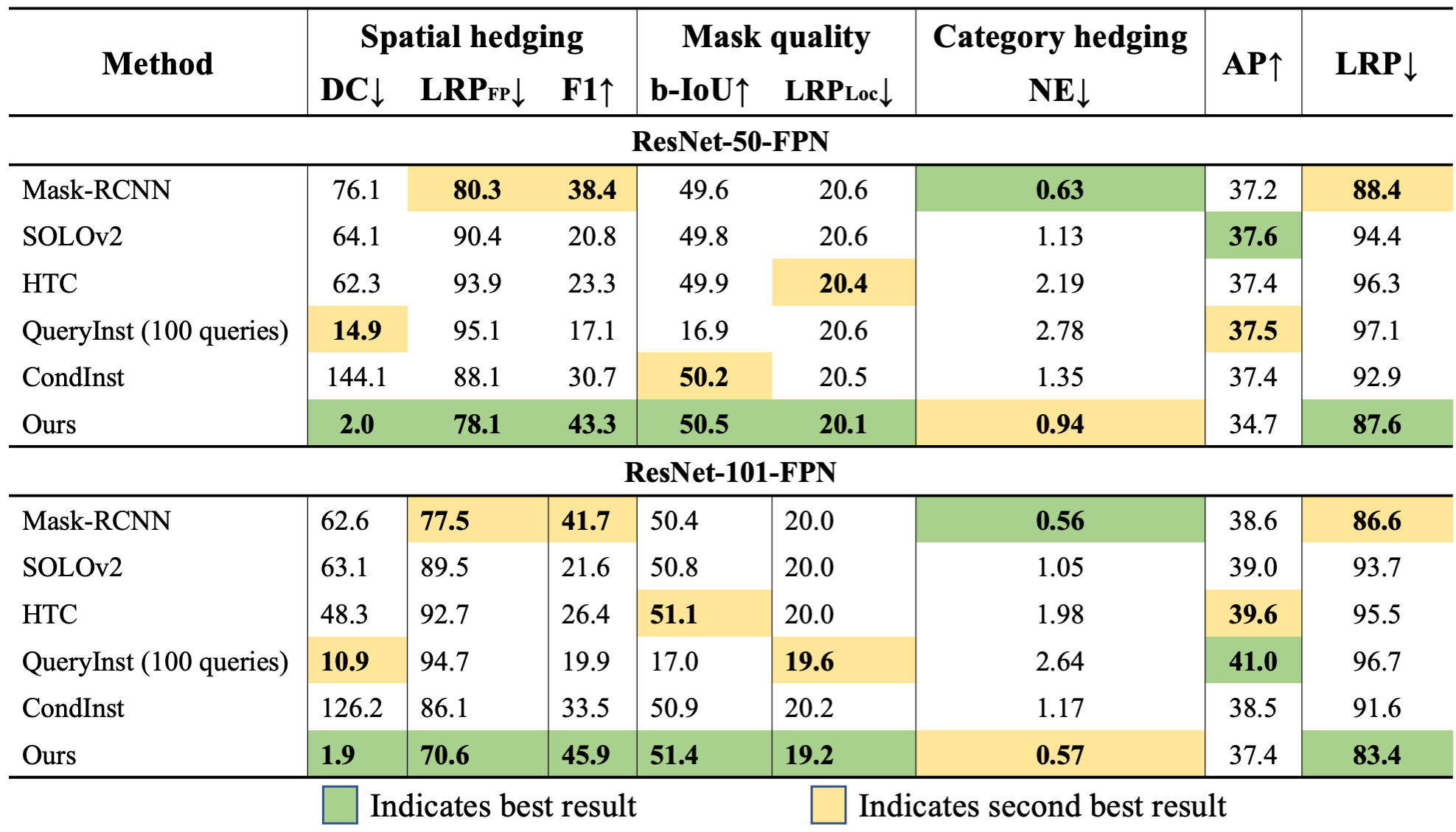}
\caption{
\textbf{Instance Segmentation on coco-val-2017}: 
Top and bottom rows are with ResNet-50-FPN and ResNet-101-FPN backbone respectively. 
Our method outperforms SOLOv2 on the F1 score by a factor of $2.08$ and performs substantially better at all metrics, showing better resolution of spatial and category hedging, without compromising mask quality.
%
} 
\label{tab:cocoresult} 
\end{table*} 

%% file: sec/5_conclusions.tex
\section{Conclusion}
Average Precision has been a longstanding metric to evaluate instance detection and segmentation.
However, AP has a few shortcomings.
We show that AP does not penalize false positives near the tail-end of the precision-recall curve.
Modern segmentation networks perform very competitively in terms of AP, but also introduce \textit{hedged predictions} which might be undesirable for a user.
We review alternate metrics in the literature, and propose measures to explicitly quantify hedging.
Modern segmentation networks turn out to have a considerable hedging problem, both spatial and categorical.
To mitigate this, we also propose a semantic sorting and NMS module. 
Experiments on three datasets show that our method considerably prunes out hedged predictions without sacrificing mask quality, while being much faster than MaskNMS.